# Visualizing and Understanding Vision System

Feng Qi 1,2 *
1. Alibaba Group
Beijing 100109
qianlong.qf@alibaba-inc.org

Guanjun Jiang 1
2. University of OXford
UK OX39DU
guanj.jianggj@alibaba-inc.com

## Abstract

*How the human vision system addresses the object identity-preserving recognition problem is largely unknown. Here, we use a vision recognition-reconstruction network (RRN) to investigate the development, recognition, learning and forgetting mechanisms, and achieve similar characteristics to electrophysiological measurements in monkeys. First, in network development study, the RRN also experiences critical developmental stages characterized by specificities in neuron types, synapse and activation patterns, and visual task performance from the early stage of coarse salience map recognition to mature stage of fine structure recognition. In digit recognition study, we witness that the RRN could maintain object invariance representation under various viewing conditions by coordinated adjustment of responses of population neurons. And such concerted population responses contained untangled object identity and properties information that could be accurately extracted via high-level cortices or even a simple weighted summation decoder. In the learning and forgetting study, novel structure recognition is implemented by adjusting entire synapses in low magnitude while pattern specificities of original synaptic connectivity are preserved, which guaranteed a learning process without disrupting the existing functionalities. This work benefits the understanding of the human visual processing mechanism and the development of human-like machine intelligence.*

## 1. Introduction

Computer vision can achieve human-level object recognition through CNN. For example, the faster region based convolutional neural network (FrRCNN) takes full spatial convolutional features map as input, and computes multiple (3000 per image) regions of interest (ROI) via its region proposal network (RPN) and categorize object in each ROI via its classifier network [30]. The human vision system is different from FrRCNN. First, human vision does not have convolutional kernel scanning across visual cells of the retina, so only object image falling in the center of the retina, i.e. fovea, can be seen clearly and processed precisely. Instead, in order to see spatially separated objects, a saccade mechanism is needed with the frontal eye field (FEF) cortex controlling sphincter and rectus muscles to zoom and move pupil rapidly to collect objects of interest in different regions [31]. Second, the human vision system supports imagination [28, 29]. For example, you can roughly imagine what you have seen or creatively imagine that a box and four wheels to form a car with top-down control signals. However, FrRCNN vision systems only have a forward processing mechanism [14, 30]. Third, the human vision system is unsupervisedly trained, so that even the deaf person can distinguish objects with a complete vision system. However, FrRCNN, etc. vision systems [32, 39] are mostly supervisedly trained, where labels are used to guide network to form proper non-linear transformation of pictures for accurate categorization. Fourth, the human cortical system including the vision system uses a population coding mechanism to represent objects or events, and to do computing and reasoning, so damage of a certain proportion of neurons will not influence normal task performance. However, units in CNN represent different features, so, for example, if the 'dog' unit of the softmax layer cannot work, FrRCNN will never articulate dog anymore. Recently, Qi [28, 29] used a fully connected visual autoencoder system to process visual information, which not only solved all the above problems, but also successfully achieved diverse tasks such as object recognition, sentence comprehension, motion judgment, language guided imagination, iterative thinking, and one-trial digit tracing, etc. on one HGLP network. Here, we implement a similar visual processing system consists of recognition and reconstruction networks (Fig.1f-g, and supplementary Fig.1) to investigate how such a fully connected visual system develops and processes visual information to support high hierarchical functions.

## 2. Development

As primates have distinct development stages (e.g. baby, adolescence and adult) characterized with significant differ-ence in neural anatomy, physiology and cognition [1], the RRN also experienced similar development process (i.e. the Early, Middle, and Late stages) characterized by 9 critical time points (ctp) each with unique firing pattern, synaptic connectivity pattern, and recognition performance (Fig.1). In the Early stage (ctp0-4), RRN took relatively few steps, 100k only, to learn to capture the salience map of objects in the retina (Fig.1b cpt0-4). The RRN started with random assigned synaptic weights (Fig.1d ctp0) that possessed no power for recognition though digits were presented in the retina (Fig.1b ctp0). Despite there was only slight synaptic plasticity in the initial stage (Fig.1d ctp0-1), the RRN started to realize that those digits were likely to appear in the center of the retina (Fig.1b cpt1). The initial activity changes occurred mainly in the last layer V1' (Fig.1c ctp0-1), indicating the importance of early development of imagination reconstruction in the primary visual cortex. The ctp2 witnessed a sudden drop in the number of excitatory synapses, so was the increase of inhibitive synapses (Fig.1a brown arrow and supplementary table1), and ctp2 is also the starting point of strength growth of both excitatory and inhibitory synapses. These changes allowed the RRN to identify whether the digits were presented on the left or right part of the retina (Fig.1b cpt2), which was clearly reflected by the horizontal segmentation of positive and negative synapses (Fig1.e). Following the same trend of synaptic alterations, the RRN could identify the vertical position and orientation of objects (Fig.1b,e ctp3), with specific cluster-like distributions gradually formed. The units also showed remarkable firing specificity with respect to digit's retinal position. In ctp4, the RRN could capture the object scale information (Fig.1b cpt4), and this was also the start point when RRN came to identify digit's morphology, namely, entering the Middle stage (ctp4-6). In the Middle stage, the most striking changes occurred in ctp5, where both excitatory and inhibitory synapses stopped strengthening and began to shrink to enhance the recognition (Fig.1a pink arrow and supplementary table1), quite similar to the pruning mechanism in the animal brain to improve the neural network efficiency [5]. At ctp6 the prune effect was quite clearly demonstrated in both activity maps with significantly sparser firing patterns (Fig.1c cpt6) and synaptic weight maps with fewer but sharper distributions (Fig.1d,e cpt6). At the same time, V1-like units formed specific visual field in retinal region (Fig.1d cpt6) [34]. From ctp6, the RRN could basically recognize and reconstruct a correct digit (Fig.1b cpt6), and the further refinement of the structural details was developed in the Late stage (ctp6-8) on a longer time scale. The excitatory to inhibitory ratio of synapse number continually declined at a moderate speed till cpt7 (Fig.1a purple arrow and supplementary table1) where the network's recognition performance reached a peak level [38, 23]. And the subsequent synaptic plasticity (in ctp8 and further) appeared to stabilize and so were other parameters. The entire process matches the development of animal and human [27, 21].

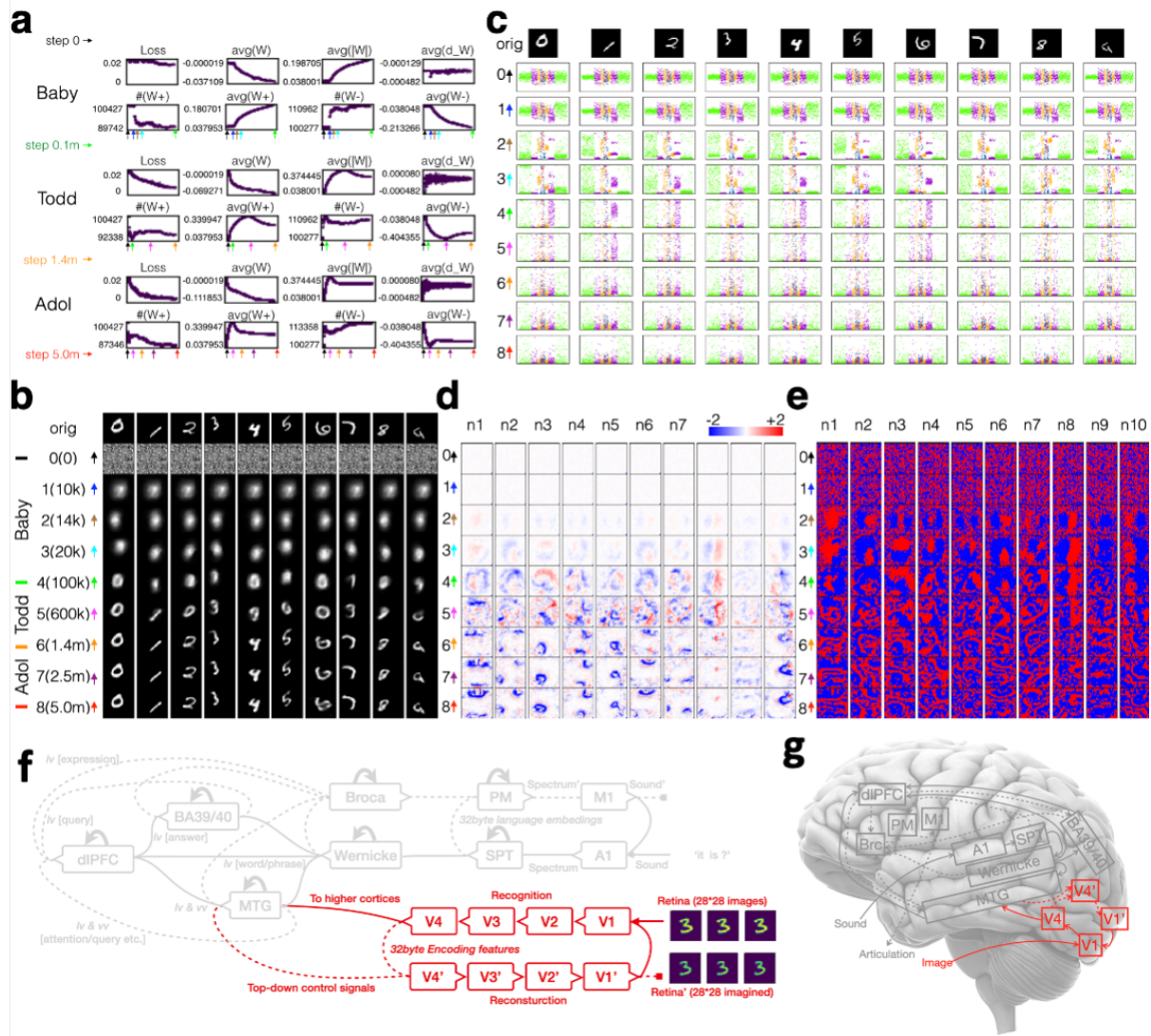

Figure 1. Development process. a, The development process contains three stages (the early, middle and late stage), and 9 critical time points (ctp) according to the key changes in synaptic weights, neuronal firing patterns, and digit reconstruction performance. b, Digit reconstructions across the development process, where RRN first figured out the digit's salience map with correct retinal location, size and orientation (early stage), then engraved the basic structure based on the salience map (middle stage), finally refined the structural details in a longer duration (late stage). c, Evolution of neuronal response maps for 10 different digit instances, ranging from initial non-specific firing to final object-related firing. Neurons are presented in order, and layers are indicated by colors. d, Evolution of synaptic weights from retina to the first layer V1 (Supplementary Fig.2 for all neurons). The increase of weight strength of both excitatory (positive) and inhibitive (negative) synapses facilitated to capture the basic characteristics of digits (ctp0-5), whereas the subsequent drop of weight strength, while keeping synaptic pattern sharpen, enabled to recognize structure details (ctp5-8). e. Evolution of the distribution of excitatory (red) and inhibitive (blue) synapses (Supplementary Fig.3 for all neurons). In the beginning, it shows random distributed patterns (ctp0-1), followed by cluster-like patterns (ctp2-4), and finally stripe-like patterns(ctp5-8). f. The visual autoencoder in the HGLP network [28]. The recognition modules (or visual encoder V1-V4) can extract the 32byte visual features from images viewed. The 32byte visual features can also be passed to high hierarchical modules such as middle temporal gyrus (MTG) for future processing such as naming the object. Moreover, 32byte visual features either come from the recognition modules or top-down control signals that can be reconstructed into images on retina' via Reconstruction modules (or visual decoder V4'-V1'). g. The brain anatomy and module connection of the visual autoencoder. Note, k represents 1,000 steps and m represents 1,000,000 steps.

## 3. Recognition

After 5 million steps of training, the RRN was mature in recognition of both digit properties and structure details. Then, we quantitatively characterized the information represented in the pivot Encoding layer (V4) to investigate how the digit properties were encoded by these neurons (n = 32). Fig. 2a shows the correlation analysis between retinal position x and neuronal activity across 128 horizontal translation trials (Supplementary Method). We witness that all layers contain certain number of x-related ($P < 0.01$) neurons, which is reasonable and even necessary, because precise digit reconstruction requires retinal position information being carried throughout the entire network. 5 out of 32 neurons in the Encoding layer are x-neurons in Fig. 2b-c, in which activity of neuron 24 (n24, the most significantly related) demonstrates three characteristics: (1) digit position modulated activity of n24; (2) the modulation relied on digit identity and morphology; (3) continuous position alteration resulted in continuous change of neuronal response. The characteristics of diffeomorphic mapping of RRN can guarantee the tolerance to position variation. All these are consistent with the characteristics of property preserving representation of inferior temporal (IT) neurons [9, 10] in monkey studies.

Similar correlation studies were performed on properties of y, s and r (Supplementary Fig. 4-7), and the 32 neurons were functionally categorized into 5 classes (Fig. 2d), which demonstrate the population coding nature [24] of these neurons. To test whether these population responses represent the digit properties, a simple linear summation decoder of downstream circuit (Supplementary Fig. 1) was implemented (Supplementary Method) to predict position x from population responses of the Encoding neurons. Fig. 2e demonstrates that the downstream layer could perceive the objective x value with a simple linear circuit. Fig. 2f shows the predictive performance in terms of R2 and mean square error (MSE), demonstrating that perceptions of retinal position x and y were most accurate, while that of digit size was worse, and in-plane orientation was the worst. Interestingly, although we didn't train the RRN with any digit properties, it could automatically untangle the manifolds of these properties [26, 8].

To investigate what individual neuronal response represents, we modulated each neuronal activity while traced the alteration of image reconstruction. Fig. 2g demonstrates how n24 response influences reconstruction of digit 5 in various x positions, where we have three observations: (1) for each column, though n24 significantly changed activity, reconstructions of digit 5 were always placed in correct x positions, which suggests that neighbor neuronal activities together also contained the position x information, and n24 activity alone could not move reconstruction position. (2) For each column, different n24 activity did severely change the reconstructed morphology, however, without communicating with neighbor neurons in the same layer, n24 could readily obtain accurate activity from feedforward inputs of previous layer, which suggests the response of n24 (the most significant x-related neuron) was induced by features in specific retinal regions. (3) across all blocks, there was no sudden change occurring no matter slight move of digit or change of n24 activity, which suggested the RRN reconstruction possessed noise tolerance to digit translation and neuronal activity [3]. Quite similar is the influence of n31 activity on digit's y position, but the reconstruction quality of morphology was more sensitive to the variation of n31 activity. Fig. 2i-j shows that both digit scaling and rotating were significantly correlated to n27 activity, which maintained fairly low but extremely important activity magnitude around 0.1, and disturbing such magnitude would substantially worsen the reconstruction quality, which suggests both big and small responses are necessary parts of the population coding [16]. In sum, in the population coding scheme, what exactly a single neuronal activity represents can hardly be described, because the interpretation may dramatically alter when neighboring neurons change firing patterns.

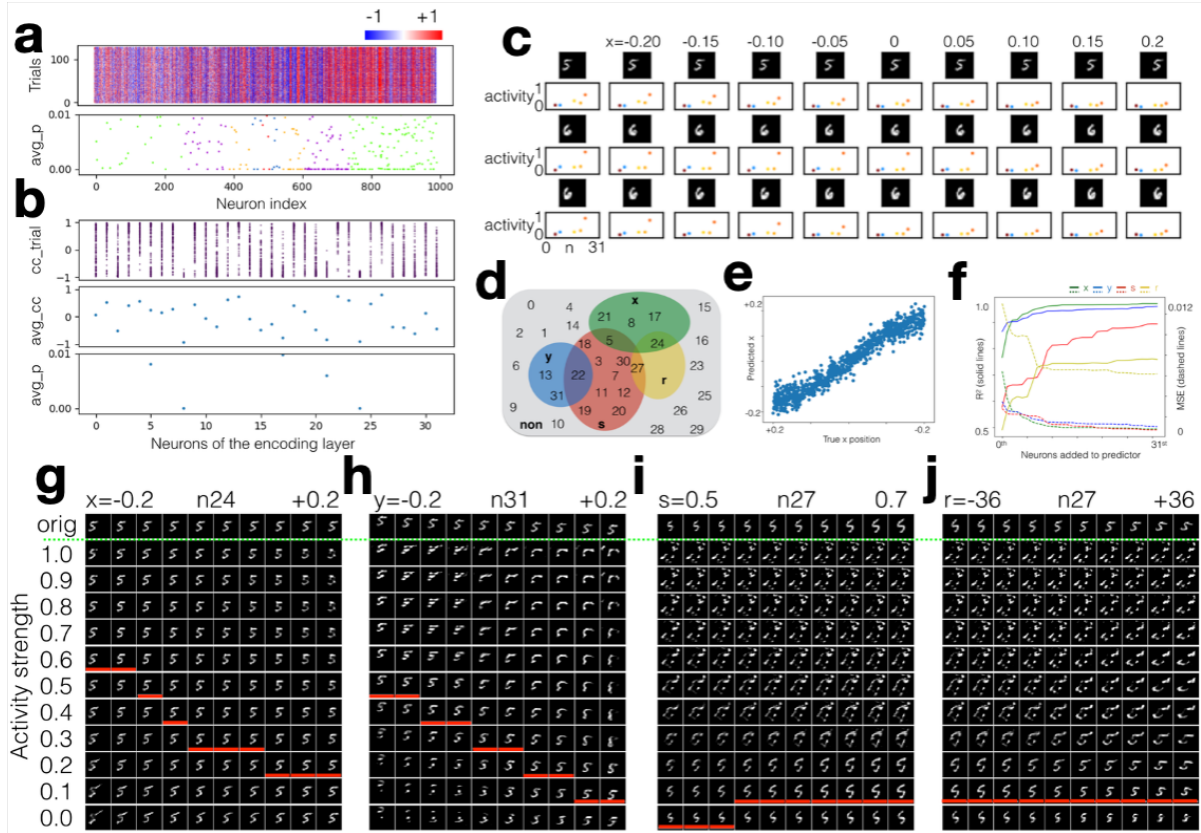

Figure 2. Neuronal responses to digit translation, scaling, and in-plane rotation. a. Neuronal responses to 128 trials of horizontal translation x (top), and significantly correlated x-neurons (bottom). Note that all layers contained certain x-neurons. b. The correlation coefficient (cc) between neuronal activity and x position for 128 trials (top), where each point represents one trial, and only 32 neurons in the Encoding layer (V4) are presented. The middle and bottom panels show the averaged cc across 128 trials and 5 significant (P < 0.01) x-neurons, respectively. c. x translation from left to right of different digit instances and corresponding responses of x-neurons. Note, activities of neuron 24 (n24, orange point, the most significantly x-related) were strongly modulated by digits' x positions, identities, and even morphologies. d. Neuronal categorization according to digits' properties. The non-specific and x, y, s, r specific neurons took a proportion of 46.9%, 15.6%, 9.4%, 34.4%, 6.3% respectively with certain overlaps. e. Prediction of position x. The encoding layer contained untangled x information that was easily extracted by a simple linear decoder implemented by a downstream layer (Supplementary Fig.1). Similar studies were performed for vertical translation (y), scaling (s), and rotation (r) in Supplementary Fig. 4-7. f. Prediction performance (coefficient of determination R2 and mean square error) for digit properties. The retinal position (x, y) were predicted with relatively high accuracy, while scaling and orientation (r, s) were with relatively low accuracy, primarily due to the innate property variations in the training set. g. The influence of n24 on digit reconstructions regarding x position. From left column to right, x position increases from -0.2 to +0.2 as illustrated on top row of original digit images. All other rows are images reconstructed with modified neuronal activities of n24. Red bars highlight the optimal activities derived from RRN. h-j similar plots for y, s, and r. Neurons were selected by lowest p-value during correlation studies.

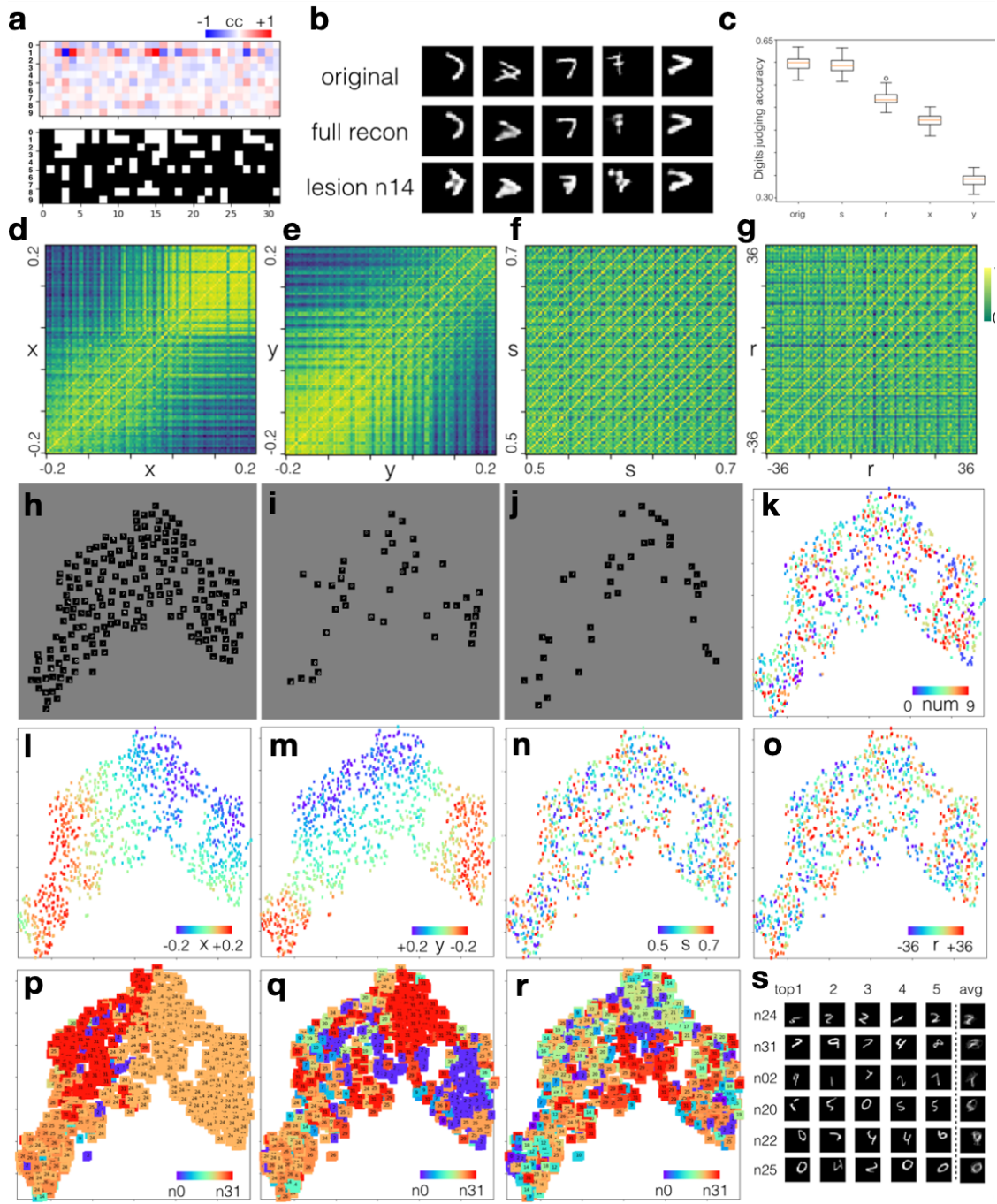

Figure 3. Digit identity recognition. a, Correlation coefficient matrix between neuronal activity and digit identity, and significantly (P <0.01) identity related neurons highlighted in white. b, Lesion of n14 (whose activity was highly correlated to the occurrence of digit 7) damaged the reconstruction of various digit 7 to different degrees. c, Digits judging accuracy without or with perturbation of x, y, s, r properties. Though training was conducted with no digit labels, a simple logistic classifier could be used to predict digit identity from population responses of the Encoding layer (Supplementary Fig.1). Note, the accuracy of natural guess is 10%. d, Population similarity matrices [12] for x translation. The 100 by 100 correlation coefficient matrix was computed between population responses across 32 Encoding neurons to each of 100 stimuli, which were constituted by ten x-position blocks, each contained ten digits from 0 to 9 (Supplementary Method). Note that the paradiagonal stripes indicate property-invariant digit selectivity. e-f are similar matrices for properties of y, s, and r. h, Non-overlapping view of digit instances in 2D tSNE space. It clearly demonstrates that these images were primarily arranged according to retinal positions, while digit identities were intertwined. i-j, Distribution of digit 0 and 1 in 2D tSNE space, respectively. k-o, Distribution of image identities and properties of x, y, s, r in 2D tSNE space. Retinal position x and y clearly dominated the global distribution of images, while digit identity, scaling and orientation showed different locally clustering characteristics. p-r, The first 3 highest responsive neurons for each image in 2D tSNE space. n24 and n31 fired strongest for most digit images, which seemed to encode features in specific retinal regions, while other neurons managed to add or subtract structural details in these regions. s, top5 favorite and average (of top20) images. These neurons fired in favor of some features in specific regions, not necessarily pointing to specific digit identity.

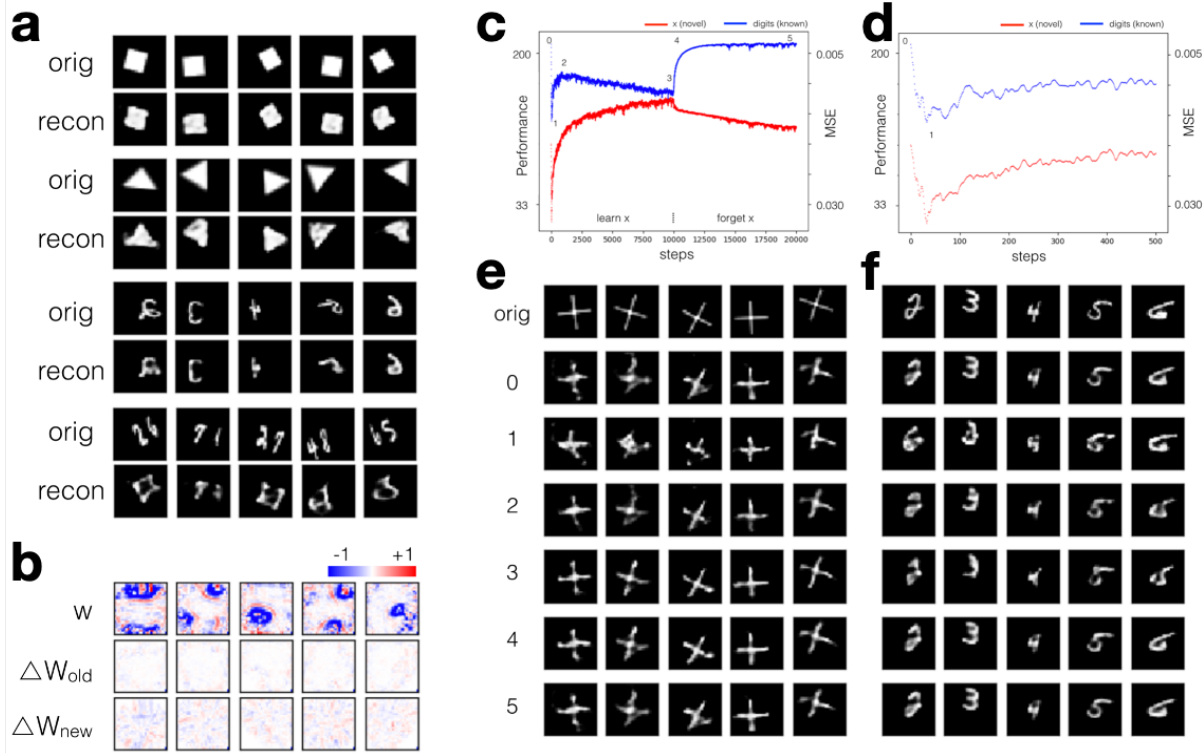

Figure 4. Learning and forgetting of the vision system. a, Reconstruction of novel structures including solid squares, triangles, mirrored digits and double digits. b, Synaptic weights distribution from the retina to V1 (first row), the alteration of synaptic weight distribution after 10,000 steps training of known structures (e.g. different digits, middle row), and weight alteration after 10,000 steps training of novel structures (e.g. symbol x, bottom row). c, Performance evolution during training and forgetting processes. The RRN was trained with novel symbol x for the first 10,000 steps, followed by known digits in the next 10,000 steps. Red represents reconstruction performance (or MSE in the upside-down y-axis) of novel structure x, while blue represents known digits. Six distinguished ctps (0-5) were observed across the entire process. d, enlarged view of the first 500 steps which clearly demonstrates the initial drop and subsequent rise of reconstruction performance. e-f, Illustrate the reconstruction of symbol x and digits for 6 ctps, which reflect the performance evolution during the learning and forgetting process.

To tackle whether neuronal activities encode digit identity, correlation coefficients between neuronal activities and digit identities were computed (Fig.3a), where nearly all neurons were involved in identity coding but biasing different digits [2, 36]. The Supplementary Fig.8 shows that though variations existed in digit morphologies, these corresponding digit-related neurons had conservative response pattern regarding each identity [37, 18]. Nevertheless, there was no single identity neuron found in the Encoding layer. For instance, n14 (the neuron most relevant to digit-7) showed diverse influence on the reconstruction of digit 7 in different conditions (Fig. 3b). Then, we used a simple logistic regression layer (Supplementary Fig.1) to assess whether neurons ensemble contains explicit digit identity information. As compared with the baseline accuracy of 10%, this simple logistic regression layer could predict digit with an accuracy of 60% (Fig.3c), though no labels were provided during the training process, which indicates that the Encoding neurons are capable of untangling digit identity which allows downstream layers to easily associate a category to such activation pattern, as the naming task in [28, 29]. In addition, we found that if the digits were placed in the retinal center, certain rotating or scaling would not drastically deteriorate the recognition accuracy, however, placing digits to peripheral regions especially on top or bottom of retina would significantly damage the digit recognition (Fig.3c) [7, 20], that's why oculomotor control is the basic function of animal neural network.

Visual processing needs to handle identity preserving image transformation problem, or to tolerate property variations during identity recognition. To investigate this, we introduced similarity matrices [12] between population responses to digits with gradual changing properties (Supplementary Method). One remarkable feature in Fig.3d is the presence of paradiagonal stripes, which indicates the same digit with decent horizontal displacement will not severely disturb the population response pattern. Another notable feature is the high correlation region on top right, which indicates that digits placed on the right part of retina will elicit similar neuronal activities regardless their identities. Vertical variation of digit position has resembled similarity matrix (Fig.3e). Nevertheless, the variation of digit size demonstrates remarkably more and stronger paradiagonal stripes than others, which suggests that for scaling effect, digits with the same morphology were encoded by similar population response pattern, and only slight modulation on population activities was needed for recognizing the size alteration (Fig.3f) [4]. In contrast, relatively large adjustment of coordinated population response was needed to recognize retinal position changes [6], which is consistent with Supplementary Fig. 9. So the natural neural mechanism in handling translational invariance is not based on max pooling [25] but on coordinated adjustment of population responses.

t-Distributed Stochastic Neighbor Embedding (tSNE) [17] could provide visible insights into how these digit instances organized in a 2D state space. Fig. 3h-j illustrate two main findings: (1) retinal position x and y dominated the global distribution of digit in the tSNE space; (2) digit identities had a tendency to cluster locally. So alteration of retinal position meant a big move in tSNE space, which required a large scale adjustment of coordinated responses to achieve it; while alteration of digit size meant a small nudge (due to unchanged x and y) in the tSNE space, which required little adjustment in population responses and showed remarkably more and stronger paradiagonal strips (Fig.3f). From Fig.3p-r, we witness that activities of n24, n31, n2 and n25 might provide location guidance information for image reconstruction, while other neurons provided relatively small activity to refine the structure in such regions. However, both big and small responses were integral parts of population coding. Fig. 3s (Supplementary Fig. 10) shows that neurons did fire in favor of features [35] in a specific retinal region, instead of digit identities which may be represented by high hierarchical modules such as the middle temporal gyri (Supplementary Fig.1).

## 4. Learning and Forgetting

Learning new and forgetting old skills (or knowledge) are basic characteristics of animal brains [19, 15, 11]. Here, we will reveal how new recognition skills are added into the working network while previous skills are still preserved without big disruption. It is not surprising, when novel structures are fed into RRN, it treats the input as known digits and reconstructs output accordingly. In Fig.4a, the RRN treated the square as compacted digit 8, and double-digit as a wide single digit. Fig.4b demonstrates that novel structures (symbol x) induced significantly (2.5 times, P = 2.43e-50) stronger synaptic plasticity than known structures (digits). More importantly, these changes occurred on all synapses in relatively small magnitude, so that previous sharp synaptic patterns could be retained [22, 33]. This guarantees that those known structures could still elicit similar neuronal responses, which enable downstream layers to process data properly and robustly without big modulation. This also guarantees that any novel structure is learned and recognized by a population coding scheme.

The population rewiring process of learning and forgetting involved several key stages (Fig. 4c-f). At ctp0, the RRN could recognize and reconstruct both novel and known structures with relatively good quality (Fig.4e,f ctp0). Surprisingly, the first 30 steps training of symbol x made a sudden drop of performance for both structures until ctp1 (Fig.4e,f ctp0-1). Novel experience made RRN unstable, but after transition ctp0-1, the RRN found a new attractor to settle on in the state space [13], thenceforth a new round of adjustment started. In transition ctp1-2, RRN rapidly improved the performance for symbol x because it was the only input to the network. Unexpectedly, the performance for known digits also got improved, which might be due to the synaptic adjustment for better reuse of previous connectivity. After sufficient recovery of RRN capacity, performance for known structure began to decline again until ctp3. After that, the RRN began to see only digits again, the network soon recovered the digit recognition performance to the original level, while forgetting of symbol x took effect rapidly (Fig.4e,f ctp3-4). For transition ctp4-5 and further, the input of digits continued to enhance the digit recognition while decrease that of symbol x. Note that the real neural network is much larger, the performance alteration (blue curve from ctp0 to ctp3) would not change such dramatically. Here, we want to differentiate between recognition learning and knowledge learning, where the former (like tasting spicy will gradually lower the perception of spicy taste) is a long term process requiring synaptic plasticity of sensory cortices, while the later (like learning a new foreign word) could be a short term process requiring certain new association connectivity formed in higher association cortices.

## 5. Discussion and Conclusion

Modern convolutional neural network (CNN) technique makes computer vision surpass human beings in tasks such as object recognition. However, computer vision still lacks intelligence in high-level cognitive functions such as visual event comprehension, description and imagination, etc. CNN first acquires images with high-resolution details, and then extracts spatial features with convolutional kernels scanning across all pixels. However, the human vision system can only effectively process objects near the fovea and needs to rapidly saccade and zoom the pupil to collect objects in different regions. Since there are essential differences in network structure and processing mechanism between the two systems, the features extracted in CNN usually can hardly find counterparts in human vision cortices, and the features required by human cognitive cortices are usually not the features given by CNN.

On the contrary, we believe the RRN network is the architecture that the human brain can achieve. Though the RRN learning and visual processing mechanism are hard to understand, the extracted features can be more easily and robustly used by high hierarchical modules. For example, RRN is not trained for classifying objects, but for reconstructing image viewed with high fidelity. Therefore, the features extracted by the encoding layer will not lose the information beyond the labels. Also, since it uses an unsupervised learning mechanism to train the model, even deaf or animals without language ability can develop an excellent RRN-like visual system. Moreover, RRN is equipped with corresponding visual decoder modules (V4'-V1'), so it has the potential to do language guided imagination. Qi's work [29] demonstrates one example of 'mental rotating' with the RRN-like visual system. In addition, the RRN uses a population coding mechanism for visual processing. As is shown above, nearly all object features, such as identity, size, morphology, can influence neuronal activities (i.e. neuronal activity does not represent single feature), and any object feature is not represented by the activity of a single neuron. Under such coding mechanism, it is difficult for us to understand what each neuron discharge represents, but the advantage is that loss of a certain number of neurons will not affect normal visual functions because the visual information is also stored in the remaining neurons. For example, in Qi's work [28], even silencing 10% of Broca neurons, the system can still articulate readable sentences based on the remaining neuronal activation.

In this paper, we also demonstrate that RRN is human brain achievable architecture from many perspectives. The RRN experiences a human-like development process from initial salience map recognition to final fine structure detection. In the object recognition task, we find that the RRN maintains object invariance representation under various conditions such as objects or head translation, rotation, and scaling, via concerted adjustment of population neuronal activities, which provides valuable insight into the study of the human vision system. Furthermore, from the study of RRN, we understand how to acquire new sensory skills while preserve old ones through adjustment of synaptic patterns. The sensory skill learning (long term unsupervised learning) uses a different mechanism from association learning (short term supervised learning), which gives us insight on how to build a human-like network with hierarchical modular modules. Under the same principle of minimizing the free energy of system [13], humans and RRN independently evolved similar neural networks, with matched observations in developments, neuronal response, and learning characteristics, etc. The work provides insights into studies of both neuroscience and machine intelligence.